\documentclass[runningheads]{llncs}
\usepackage{graphicx}
\usepackage{amsmath,amssymb} % define this before the line numbering.
\usepackage{color}
\usepackage[width=122mm,left=12mm,paperwidth=146mm,height=193mm,top=12mm,paperheight=217mm]{geometry}
\begin{document}
% \renewcommand\thelinenumber{\color[rgb]{0.2,0.5,0.8}\normalfont\sffamily\scriptsize\arabic{linenumber}\color[rgb]{0,0,0}}
% \renewcommand\makeLineNumber {\hss\thelinenumber\ \hspace{6mm} \rlap{\hskip\textwidth\ \hspace{6.5mm}\thelinenumber}}
% \linenumbers
\pagestyle{headings}
\mainmatter

\newcommand{\Vector}[1]{\lowercase{\mathbf{#1}}}
\newcommand{\VectorArray}[1]{\left(#1\right)}
\newcommand{\Matrix}[1]{\uppercase{\mathsf{#1}}}
\newcommand{\SizedMatrix}[3]{\Matrix{#1}_{#2\times#3}}
\newcommand{\MatrixArray}[1]{\left(#1\right)}
\newcommand{\Skew}[1]{[#1]_\times}
\newcommand{\Transpose}[1]{#1^\top}
\newcommand{\Equation}[1]{{\langle}#1{\rangle}}
\newcommand{\Nister}{Nist\'{e}r}
\newcommand{\Stewenius}{Stew\'{e}nius}
\newcommand{\ApproxR}{\SizedMatrix{I}{3}{3}+\Skew{\Vector{r}}}

\newcommand{\Groebner}{Gr\"{o}bner}

\newcommand{\Spherical}{\textbf{Spherical}}
\newcommand{\SphericalPoly}{\textbf{SphericalPoly}}
\newcommand{\Stew}{\textbf{Stew\'{e}nius}}

\title{Structure from motion on a sphere} % Replace with your title

\titlerunning{Structure from motion on a sphere}

\authorrunning{Jonathan Ventura}

\author{Jonathan Ventura}

%Please write out author names in full in the paper, i.e. full given and family names. 
%If any authors have names that can be parsed into FirstName LastName in multiple ways, please include the correct parsing, in a comment to the volume editors:
%\index{Lastnames, Firstnames}
%(Do not uncomment it, because you may introduce extra index items if you do that...)

\institute{Department of Computer Science,\\
	University of Colorado Colorado Springs\\
	\email{ jventura@uccs.edu}
}

\maketitle

\begin{abstract}
We describe a special case of structure from motion where the camera rotates on a sphere.  The camera's optical axis lies perpendicular to the sphere's surface.  In this case, the camera's pose is minimally represented by three rotation parameters.  From analysis of the epipolar geometry we derive a novel and efficient solution for the essential matrix relating two images, requiring only three point correspondences in the minimal case.  We apply this solver in a structure-from-motion pipeline that aggregates pairwise relations by rotation averaging followed by bundle adjustment with an inverse depth parameterization.  Our methods enable scene modeling with an outward-facing camera and object scanning with an inward-facing camera.
\end{abstract}

\section{Introduction}

Accurate visual 3D reconstruction is highly dependent on establishing sufficient baseline between images so that the translation between them can be reliably estimated and 3D points can be accurately triangulated.  However, we have found that, in practice, it is difficult for an untrained user to capture image sequences with sufficient baseline; typically, the natural inclination is to rotate the camera instead of translating it, which causes the structure-from-motion system to fail.

In this work, we instead specifically target camera rotation as the basis for structure from motion.  The critical assumption here is that the camera rotates at some fixed distance from the origin, with its optical axis aligned with the ray between the origin and the camera center.  We call this ``spherical motion.'' 

The camera could be pointing inward or outward.  An example of an inward-facing camera would be object scanning setups such as a turntable or spherical gantry.  An example of an outward-facing camera would be a typical user capturing a panorama -- the user holds the camera away from their body at a fixed distance while rotating.

In either case, the global scale of the 3d reconstruction is unknown, as is always the case in pure monocular structure-from-motion.  The global scale is determined by the radius of the sphere, which we arbitrarily set to unit length.  However, what is interesting about this particular case of camera motion is that the relative scale between camera pairs is known, because the radius of the sphere is fixed.  This is a distinct advantage over general monocular camera motion estimation, where the relative scale of the translation between camera pairs must be determined by point triangulation and scale propagation, which is highly susceptible to scale drift.  With spherical camera motion, we can directly compose relative pose estimates to determine the complete camera trajectory without needing to propagate scale.  A second advantage is that the relative pose between cameras is fully determined by three rotational degrees of freedom and so can be estimated from three point correspondences as opposed to the five correspondences needed in the general case.

In this paper, after a survey of related work (Section \ref{sec:related_work}), we analyze the geometry of spherical camera motion (Section \ref{sec:geometry}) and derive efficient solvers for the essential matrix (Section \ref{sec:solution}).  We integrate these solvers into a complete structure-from-motion pipeline (Section \ref{sec:pipeline}) and present an evaluation of our methods on synthetic and real data (Section \ref{sec:evaluation}) followed by conclusions and future work (Section \ref{sec:conclusions}).

\section{Related Work}
\label{sec:related_work}

A particular problem of interest in geometric computer vision is inferring the essential matrix relating two images from point correspondences, especially from a minimal set of correspondences \cite{longuet1987computer}.  Minimal solutions are useful for application in a random sample consensus (RANSAC) \cite{Fischler:1981:RSC:358669.358692} loop to robustly estimate the motion parameters and separate inliers from outliers.  \Nister~\cite{1288525} derived an efficient minimal solution from five point correspondences and \Stewenius~et al.~\cite{Stewénius2006284} later improved the accuracy of this method.  In this work, we derive a solution for the essential matrix from at least three correspondences which applies when the camera undergoes spherical motion.

Several previous works have considered solutions for monocular relative pose given circular motion or single-axis rotation as observed with a turntable \cite{Fitzgibbon:1998uh,Mendonca:2001ha,Jiang:2004ej} or a non-holonomic vehicle \cite{5152255}.  In this work we derive a spherical motion solver which allows three rotational degrees of freedom and thus requires three point correspondences in the minimal case.

Also closely related are the works by Peleg and Ben-Ezra \cite{786969} and Shum and Szeliski \cite{791191} on stereo or multi-perspective panoramas.  In these works, an outward-facing camera is spun on a circular path and images are captured at regular intervals.  They demonstrated that by careful sampling of the images, stereo cylindrical panoramas can be created and used for either surround-view stereo viewing or 3D stereo reconstruction.  These works use either controlled capture on a turntable \cite{791191} or manifold mosaicing \cite{609346} to obtain the positions of the images in the sequence, whereas we develop an automatic, accurate structure-from-motion pipeline which applies to both circular and spherical motion image sequences.

Structure-from-motion refers to simultaneously estimating camera poses and 3D scene geometry from an image sequence or collection.  For example, Pollefeys et al.~describe a pipeline for visual modeling with a handheld camera \cite{Pollefeys:2004:VMH:986694.986705}, and Snavely et al.~developed a structure-from-motion system from unstructured internet image collections \cite{snavely2006photo}.  Most related to the present work is the 1DSfM approach of Wilson and Snavely \cite{Wilson2014}, where the camera orientations are first estimated using a robust global rotation averaging approach \cite{Chatterjee_2013_ICCV} and then the camera translations are estimated separately.  In our approach, the camera's position is directly determined by its orientation, so the second translation estimation step is not needed.

\section{The Geometry of Spherical Camera Motion}
\label{sec:geometry}

In this section we give expressions for the absolute and relative pose matrices induced by spherical motion and derive the form of the essential matrix.

\subsection{Camera Extrinsics}

\subsubsection{Inward-facing Camera}
For an inward-facing camera, the $3\times4$ camera extrinsics matrix $\Matrix{P}$ can be expressed using a $3\times3$ rotation matrix $\Matrix{R}$ and a $3\times1$ vector $\Vector{z}$:
\begin{equation}
    \Matrix{P}_{\mathrm{in}} = [ \Matrix{R}~|~\Vector{z} ],
\end{equation}
where $\Vector{z} = \Transpose{[0~0~1]}$.

\subsubsection{Outward-facing Camera}
For an outward-facing camera, the translation direction is reversed:
\begin{equation}
    \Matrix{P}_{\mathrm{out}} = [ \Matrix{R}~|~-\Vector{z} ].
\end{equation}

\subsection{Relative Pose}

Given two inward-facing cameras with extrinsics $\Matrix{P}_1=[ \Matrix{R}_1~|~\Vector{z}]$ and $\Matrix{P}_2=[\Matrix{R}_2~|~\Vector{z} ]$, we can now derive the relative pose $[ \Matrix{R}~|~\Vector{t}_{\mathrm{in}}]$  between them.  The relative rotation is
\begin{equation}
\Matrix{R}=\Matrix{R}_2 \Transpose{\Matrix{R}_1}.
\end{equation}
and the relative translation is
\begin{equation}
\label{eq:tin}
    \Vector{t}_{\mathrm{in}} = \Vector{z} - \Vector{r}_3
\end{equation}
where $\Vector{r}_3$ denotes the third column of $\Matrix{R}$.

For outward-facing cameras with relative pose $[ \Matrix{R}~|~\Vector{t}_{\mathrm{out}}]$, the rotation is the same and the translation direction is reversed:
\begin{equation}
\label{eq:tout}
    \Vector{t}_{\mathrm{out}} = \Vector{r}_3 - \Vector{z}.
\end{equation}

\subsection{Essential Matrix}

The essential matrix $\Matrix{E}$ relates corresponding camera normalized (i.e.~calibrated) homogeneous points $\Vector{u}$ and $\Vector{v}$ in two images such that
\begin{equation}
    \Transpose{\Vector{v}} \Matrix{E} \Vector{u} = 0.
\end{equation}
If the two images have relative pose $[ \Matrix{R} | \Vector{t} ]$  then
\begin{equation}
    \Matrix{E} = \Skew{\Vector{t}} \Matrix{R},
\end{equation}
where $\Skew{\Vector{a}}$ is the skew-symmetric matrix such that $\Skew{\Vector{a}}\Vector{b} = \Vector{a} \times \Vector{b}~\forall~\Vector{b}$.  

Plugging in the relative pose expressions given above, the essential matrices for inward- and outward-facing cameras are
\begin{equation}
\label{eq:Ein}
    \Matrix{E}_{\mathrm{in}} = \Skew{\Vector{z} - \Vector{r}_3} \Matrix{R}
\end{equation}
and
\begin{equation}
\label{eq:Eout}
    \Matrix{E}_{\mathrm{out}} = \Skew{\Vector{r}_3 - \Vector{z}} \Matrix{R}.
\end{equation}

Note that $\Matrix{E}_{\mathrm{in}} = -\Matrix{E}_{\mathrm{out}}$.  Since the essential matrix is only defined up to scale, the essential matrix for inward- and outward-facing cameras underoing the same relative rotation is equivalent.

\section{Solving for the Essential Matrix}
\label{sec:solution}

Here we characterize the essential matrix relating cameras undergoing spherical motion and derive a solution for all possible essential matrices arising from at least three correspondences between two images.

In general, the essential matrix has five degrees of freedom \cite{hartley2003multiple}, corresponding to three rotational degrees of freedom and two translational, since in the general case the translation is only defined up to scale.  However, the special form of essential matrix for spherical motion derived above is determined completely by three rotational degrees of freedom.  This implies that we can solve for the essential matrix using only three correspondences, as opposed to the five correspondences needed in the general case \cite{1288525}.

The essential matrix for spherical motion has a special form and can be fully described by six parameters $e_1, \ldots, e_6$:
\begin{equation}
    \Matrix{E} =
    \begin{bmatrix}
    e_{1} & e_{2} & e_{3} \\
    e_{2} & -e_{1} & e_{4} \\
    e_{5} & e_{6} & 0
    \end{bmatrix}
\end{equation}
This can be derived using the fact that, since $\Matrix{R}$ is orthonormal, each column can be expressed as a cross product of the other two.

\subsection{Finding the Nullspace}
Given $n \geq 3$ corresponding image points $\Vector{u}_1,\ldots,\Vector{u}_n$ and $\Vector{v}_1,\ldots,\Vector{v}_n$, we have $n$ epipolar constraint equations of the form
\begin{equation}
        \Transpose{\Vector{v}_i} \Matrix{E} \Vector{u}_i = 0.
\end{equation}

We re-arrange and stack the epipolar constraints to form a linear system on the parameters:
\begin{equation}
\label{eq:system}
\begin{bmatrix}
u_{11}v_{11} - u_{12}v_{12} & u_{11}v_{12} + u_{12}v_{11} & u_{13}v_{11} & u_{13}v_{12} & u_{11}v_{13} & u_{12}v_{13}\\
\vdots & \vdots & \vdots & \vdots & \vdots & \vdots \\
u_{n1}v_{n1} - u_{n2}v_{n2} & u_{n1}v_{n2} + u_{n2}v_{n1} & u_{n3}v_{n1} & u_{n3}v_{n2} & u_{n1}v_{n3} & u_{n2}v_{n3}\\
\end{bmatrix}
\begin{bmatrix}
e_1 \\
e_2 \\
e_3 \\
e_4 \\
e_5 \\
e_6 
\end{bmatrix}
= \Vector{0}
\end{equation}
where $u_{ij}$ denotes the j-th element of $\Vector{u}_i$.

We now find three $6 \times 1$ vectors $\Vector{b}_1,\Vector{b}_2,\Vector{b}_3$ spanning the right nullspace of the $n \times 6$ matrix on the left-hand side of Equation \ref{eq:system}.  The essential matrix must be of the form
\begin{equation}
\label{eq:Eform}
\Matrix{E} =
\begin{bmatrix}
b_{11}x + b_{21}y + b_{31}z& b_{12}x + b_{22}y + b_{32}z& b_{13}x + b_{23}y + b_{33}z\\
b_{12}x + b_{22}y + b_{32}z& - b_{11}x - b_{21}y - b_{31}z& b_{14}x + b_{24}y + b_{34}z\\
b_{15}x + b_{25}y + b_{35}z& b_{16}x + b_{26}y + b_{36}z& 0
\end{bmatrix}
\end{equation}
for some scalars $x,y,z$.  Here $b_{ij}$ denotes the $j$-th element of vector $b_i$.

Any choice of scalars $x,y,z$ will produce a solution for $\Matrix{E}$ which satisfies the epipolar constraints.  However, a second requirement is that the matrix must satisfy the properties of an essential matrix; namely, that it is rank two and that both non-zero singular values are equal \cite{Faugeras:1993vt}.  These properties lead to non-linear constraints which are solved in the following subsection.

\subsection{Applying Non-linear Constraints}

The requirements on the singular values of the essential matrix are enforced by the following cubic constraints \cite{Faugeras:1993vt}:
\begin{equation}
    \Matrix{E}\Transpose{\Matrix{E}}\Matrix{E}-\frac{1}{2}\mathrm{trace}(\Matrix{E}\Transpose{\Matrix{E}})\Matrix{E} = \Matrix{0}.
\end{equation}
This $3 \times 3$ matrix equation gives a system of nine cubic constraints in $x,y,z$.  Since the essential matrix is only determined up to scale, we let $z=1$.

Using a symbolic math toolbox, we found that this system has rank six, and that the second and third rows of the system form a linearly independent set of six equations.

We separate these six equations into a $6 \times 10$ matrix $\Matrix{A}$ of coefficients and a vector $\Vector{m}$ of $10$ monomials such that 
\begin{equation}
\label{eq:Am}
    \Matrix{A} \Vector{m} = \Vector{0}
\end{equation}
where
\begin{equation}
    \Vector{m} =
    \Transpose{
    \begin{bmatrix}
       x^3 &
     x^2 y &
     x y^2 &
       y^3 &
       x^2 &
       x y &
       y^2 &
         x &
         y &
         1
    \end{bmatrix}
    }.
\end{equation}

\subsection{Solution using the Action Matrix Method}
\label{sec:actionmatrix}

The action matrix method has been established  as a general tool to solve systems of polynomial equations arising from geometric computer vision problems \cite{Kukelova2008}.  Briefly, once we have found a Gr\"{o}bner basis \cite{cox1992ideals,cox2006using} for the system of polynomial equations, we can derive a transformation from the coefficient matrix to an action matrix whose eigenvalues and eigenvectors contain the solutions.

Using the Macaulay2 algebraic geometry software system, we determined that Equation \ref{eq:Am} has at most four solutions. By ordering the monomials in $\Vector{m}$ using graded reverse lexicographic ordering and running Gauss-Jordan elimination on $\Matrix{A}$, we immediately arrive at a Gr\"{o}̈bner basis for the ideal $I$ generated by the six polynomial equations, since this leaves only four monomials that are not divisible by any of the leading monomials in the equations. These monomials form a basis for the quotient ring $\mathbb{C}[x, y]/I$ and are the same basis monomials reported by Macaulay2.

Let $\Matrix{G}$ be the $6 \times 4$ matrix such that $\begin{bmatrix} \Matrix{I}_6 & \Matrix{G}\end{bmatrix}$ is the result of running Gauss-Jordan elimination on $\Matrix{A}$. Now we have
\begin{equation}
    \begin{bmatrix} \Matrix{I}_6 & \Matrix{G}\end{bmatrix}
    \Vector{m} = \Vector{0}
\end{equation}
which implies that
\begin{align}
      x^3 + G_{11} y^2 + G_{12} x + G_{13} y + G_{14} &= 0\\
    x^2 y + G_{21} y^2 + G_{22} x + G_{23} y + G_{24} &= 0\\
    \label{eq:G3}
    x y^2 + G_{31} y^2 + G_{32} x + G_{33} y + G_{34} &= 0\\
      y^3 + G_{41} y^2 + G_{42} x + G_{43} y + G_{44} &= 0\\
    \label{eq:G5}
      x^2 + G_{51} y^2 + G_{52} x + G_{53} y + G_{54} &= 0\\
    \label{eq:G6}
      x y + G_{61} y^2 + G_{62} x + G_{63} y + G_{64} &= 0.
\end{align}
Using Equations \ref{eq:G3}, \ref{eq:G5} and \ref{eq:G6} we can define a $4 \times 4$ matrix $\Matrix{A}_x$ as
\begin{equation}
    \Matrix{A}_x =
    \begin{bmatrix}
    -G_{31} & -G_{32} & -G_{33} & -G_{34} \\
    -G_{51} & -G_{52} & -G_{53} & -G_{54} \\
    -G_{61} & -G_{62} & -G_{63} & -G_{64} \\
    0 & 1 & 0 & 0
    \end{bmatrix}
\end{equation}
so that 
\begin{equation}
\Matrix{A}_x 
\begin{bmatrix}
y^2 \\
x \\
y \\
1
\end{bmatrix}
=
x
\begin{bmatrix}
y^2 \\
x \\
y \\
1
\end{bmatrix}.
\end{equation}
Thus the eigenvalues of $\Matrix{A}_x$ are solutions for $x$, and the eigenvectors contain corresponding solutions for $y$.  $\Matrix{A}_x$ is the ``action matrix'' for $x$ and $y^2,x,y,1$ are the basis monomials.

Once we have found up to four real-valued solutions for $x$ and $y$ by eigendecomposition of $\Matrix{A}_x$, we apply them in Equation \ref{eq:Eform} to produce four solutions for the essential matrix $\Matrix{E}$.

\subsection{Solution by Reduction to Single Polynomial}
\label{sec:polynomial}

A possibly faster method to find solutions for $x$ and $y$ would be to use the characteristic polynomial of $\Matrix{A}_x$ to find its eigenvalues:
\begin{equation}
    |\Matrix{A}_x - x \Matrix{I}_3| = 0.
\end{equation}
This involves computing the determinant of a $4 \times 4$ matrix of polynomials in $y$.  We found that a slight speedup is possible by transforming the problem to instead use a $3 \times 3$ symbolic determinant.

First, we define $\Vector{m}'$ which is a reordering the monomials in $\Vector{m}$:
\begin{equation}
    \Vector{m}' =
    \Transpose{
    \begin{bmatrix}
   x^3 &
 x^2 y &
 x y^2 &
   y^3 &
   y^2 &
     y &
   x^2 &
   x y &
     x &
     1
    \end{bmatrix}
    }.
\end{equation}
The system of equations $\Matrix{A}\Vector{m}=\Vector{0}$ from Equation \ref{eq:Am} is rewritten using this new ordering.  We form a reordered matrix of coefficients $\Matrix{A}'$ such that
\begin{equation}
    \Matrix{A}'\Vector{m}' = \Vector{0}.
\end{equation}

Let $\Matrix{G}'$ be the $6 \times 4$ matrix such that $\begin{bmatrix} \Matrix{I}_6 & \Matrix{G}'\end{bmatrix}$ is the result of running Gauss-Jordan elimination on $\Matrix{A}'$. Now we have
\begin{equation}
    \begin{bmatrix} \Matrix{I}_6 & \Matrix{G}'\end{bmatrix}
    \Vector{m}' = \Vector{0}
\end{equation}
which implies that
\begin{align}
  x^3 + G'_{11} x^2 + G'_{12} x y + G'_{13} x + G'_{14} &= 0\\
x^2 y + G'_{21} x^2 + G'_{22} x y + G'_{23} x + G'_{24} &= 0\\
x y^2 + G'_{31} x^2 + G'_{32} x y + G'_{33} x + G'_{34} &= 0\\
\label{eq:Gp4}
  y^3 + G'_{41} x^2 + G'_{42} x y + G'_{43} x + G'_{44} &= 0\\
\label{eq:Gp5}
  y^2 + G'_{51} x^2 + G'_{52} x y + G'_{53} x + G'_{54} &= 0\\
\label{eq:Gp6}
    y + G'_{61} x^2 + G'_{62} x y + G'_{63} x + G'_{64} &= 0.
\end{align}
Using Equations \ref{eq:Gp4}, \ref{eq:Gp5} and \ref{eq:Gp6} we can define a $3 \times 3$ matrix $\Matrix{B}(y)$ as
\begingroup
\setlength\arraycolsep{5pt}
\begin{equation}
    \Matrix{B}(y) =
    \begin{bmatrix}
G'_{41} & G'_{43} + G'_{42}y & G'_{44} + y^3\\
G'_{51} & G'_{53} + G'_{52}y & G'_{54} + y^2\\
G'_{61} & G'_{63} + G'_{62}y & G'_{64} + y
    \end{bmatrix}
\end{equation}
\endgroup
so that
\begin{equation}
    \Matrix{B}(y)
    \begin{bmatrix}
    x^2 \\
    x \\
    1
    \end{bmatrix}
    = \Vector{0}.
\end{equation}
Because $\Matrix{B}(y)$ has a null vector, its determinant must be equal to zero, leading to a quartic polynomial $\Equation{n}$ in $y$:
\begin{equation}
    \Equation{n} \equiv |\Matrix{B}(y)| = 0.
\end{equation}

The quartic polynomial $\Equation{n}$ can be solved in closed-form using Ferrari's method.  Once we have four solutions for $y$, the corresponding solutions for $x$ are found by finding a null vector of $\Matrix{B}(y)$.  Then the solutions for $x$ and $y$ are used to produce solutions for the essential matrix using Equation \ref{eq:Eform}.

\subsection{Decomposition of the Essential Matrix}
\label{sec:decomposeE}

Once we have a solution for the essential matrix, we need to decompose it into a rotation and translation to find the relative pose.  The decomposition follows the normal procedure for extracting a ``twisted pair'' of solutions \cite{hartley2003multiple}, giving two solutions for the rotation, $\Matrix{R}_a$ and $\Matrix{R}_b$, and one solution for the translation direction $\hat{\Vector{t}}$ which is only determined up to scale.

Let $\Matrix{E} \sim \Matrix{U} \Matrix{S} \Transpose{\Matrix{V}}$ be the singular value decomposition of $\Matrix{E}$ where $\Matrix{U}$ and $\Matrix{V}$ are chosen such that $|\Matrix{U}| > 0$ and $|\Matrix{V}| > 0$.  Define matrix $\Matrix{D}$ as
\begin{equation}
    \Matrix{D} =
    \begin{bmatrix}
    0 & 1 & 0 \\
    -1 & 0 & 0 \\
    0 & 0 & 1
    \end{bmatrix}.
\end{equation}
Then $\Matrix{R}_a = \Matrix{U}\Matrix{D}\Transpose{\Matrix{V}}$ and $\Matrix{R}_b = \Matrix{U}\Transpose{\Matrix{D}}\Transpose{\Matrix{V}}$.  The solution for the translation direction is $\hat{\Vector{t}}=\Transpose{[ U_{13}~U_{23}~U_{33} ]}$.

Only one of the rotations is consistent with spherical motion.  Let $\Vector{t}_a$ and $\Vector{t}_b$ be corresponding translation vectors for rotation solutions $\Matrix{R}_a$ and $\Matrix{R}_b$, respectively, determined by Equation \ref{eq:tin} if the cameras are inward-facing or Equation \ref{eq:tout} if the cameras are outward-facing.  We can choose the correct relative pose solution by choosing the rotation whose corresponding translation is closest to the translation solution $\Vector{t}$.

Specifically, we define scores $s_a$ and $s_b$ according to the absolute value of the normalized dot product between $\Vector{t}_a$ or $\Vector{t}_b$ and $\hat{\Vector{t}}$:
\begin{equation}
    s_a = \frac{ | \Vector{t}_a \cdot \hat{\Vector{t}} | }{ || \Vector{t}_a || }, s_b = \frac{ | \Vector{t}_b \cdot \hat{\Vector{t}} | }{ || \Vector{t}_b || }.
\end{equation}
The solution with higher score is chosen as the correct relative pose:
\begin{equation}
    [\Matrix{R} | ~\Vector{t}] =
    \begin{cases}
    [ \Matrix{R}_a | ~\Vector{t}_a ] & \textrm{if } s_a > s_b, \\
    [ \Matrix{R}_b | ~\Vector{t}_b ] & \textrm{otherwise}.
    \end{cases}
\end{equation}

\section{An Integrated Structure from Motion Pipeline}
\label{sec:pipeline}

The algorithms described in Section \ref{sec:solution} enable us to solve for the relative pose between two cameras from a minimal or overdetermined set of image correspondences.  The minimal solver is useful for random sample consensus (RANSAC) \cite{Fischler:1981:RSC:358669.358692} where we compute relative pose hypotheses from randomly sampled minimal sets of correspondences and accept the hypothesis with the highest number of inliers.

In this section we describe an integrated structure-from-motion pipeline which uses our novel solvers to recover the camera trajectory from a spherical motion image sequence by aggregating pairwise relationships and produce a 3D point cloud reconstruction of the scene.

The input to the pipeline is a sequence of images captured by an inward- or outward-facing camera undergoing spherical motion.  We assume the camera is pre-calibrated so that its intrinsic parameters including focal length, principal point, and radial and tangential distortion coefficients are known.

\subsection{Feature Tracking and Relative Pose Estimation}

We first use feature detection and tracking to establish image correspondences between neighboring images in the sequence.  Between each successive pair of images in the sequence, we apply one of our spherical motion solvers from Section \ref{sec:solution} in a Preemptive RANSAC loop \cite{1238341} in order to robustly estimate the essential matrix and find a consensus set of inliers.  To test inliers we threshold the Sampson error \cite{hartley2003multiple} between the epipolar line and the image point in the second image.  Outlier feature tracks are removed from consideration in successive frames.

Either minimal solver gives at most four solutions for the essential matrix from three correspondences.  We choose the essential matrix with lowest error on a fourth randomly sampled correspondence.  Finally, the decomposition of the essential matrix gives two possible rotations $\Matrix{R}_a$ and $\Matrix{R}_b$ which are disambiguated using the score function described in Section \ref{sec:decomposeE}.

\subsection{Loop Closure}

After processing the images in sequence, we detect loop closures by matching features between non-neighboring images.  Each feature in image $i$ is matched to its nearest neighbor in image $j$.  The set of putative matches bewtween images $i$ and $j$ are then filtered using Preemptive RANSAC with one of our minimal solvers, and the resulting relative pose is recorded if the number of inliers exceeds a threshold.

\subsection{Global Pose Estimation by Rotation Averaging}

At this point, we have a set of estimated rotations $\Matrix{R}_{ij}$ between images $i$ and $j$ in the sequence.  Now we can apply rotation averaging \cite{Hartley:2013bb} to produce a global estimate of all camera orientations.  Specifically, we use the robust L1 rotation averaging method of Chatterjee and Govindu \cite{Chatterjee_2013_ICCV}.  Since the translation of each camera is fully determined by the camera's rotation, this effectively produces an estimate of all camera poses.

\subsection{Inverse Depth Bundle Adjustment}

Finally, we refine the camera pose estimates using bundle adjustment.  The output of the previous steps is an estimated pose $[ \Matrix{R}_i | \Vector{t_i} ]$ for each camera and features matches between images.  The feature matches are aggregated into feature tracks across multiple images.

The rotation $\Matrix{R}_i$ of each camera is parameterized by a $3 \times 1$ vector $\Vector{r}_i$ where $\Matrix{R}_i = \textrm{exp}_{SO(3)}(\Vector{r}_i)$.  Since the translation is vector is fixed, we do not need to explicitly parameterize it.

We found that, especially with an outward-facing camera, the traditional methods of algebraic triangulation and bundle adjustment over 3D point locations are unstable because of the small baselines involved.  Instead, we use an inverse depth parameterization which extends the work of Yu and Gallup \cite{Yu14}.

Each 3D point $\Vector{x}_j$ has a designated reference camera with index $n_j$ so that
\begin{equation}
    \Vector{x}_j = \Transpose{\Matrix{R}_{n_j}} ( w_j \Vector{u}_{n_j,j} - \Vector{t}_{n_j} )
\end{equation}
where $\Vector{u}_{n_j,j}$ is the observation of point $j$ in camera $n_j$ and $w_j$ is the inverse depth of point $j$.  We set the reference camera of each point to be the first camera which observed it in the image sequence.

The output of global rotation averaging gives an initialization for the camera rotations.  To initialize the 3D points, we first linearly solve for $w_j$ using all observations of the point.  Then we use non-linear optimization over all rotation and inverse depth parameters to minimize the total robustified re-projection error
\begin{equation}
    \sum_{(i,j) \in \mathcal{V}} h(|| \pi(\Vector{u}_{i,j}) - \pi(\Matrix{R}_i \Vector{x}_j + \Vector{t}_i) ||^2)
\end{equation}
where $\pi(\Vector{u})$ is the perspective projection function $\pi( \Transpose{[ x~y~z ]} ) = \Transpose{[ x/z~y/z ]}$ and $(i,j) \in \mathcal{V}$ if camera $i$ observes point $j$.  $h(\cdot)$ is the Huber cost function which robustifies the minimization against outlier measurements \cite{huber1964}.

\section{Evaluation}
\label{sec:evaluation}

\subsection{Essential Matrix Solvers}

\begin{figure}[t]
    \centering
    \includegraphics[width=6cm]{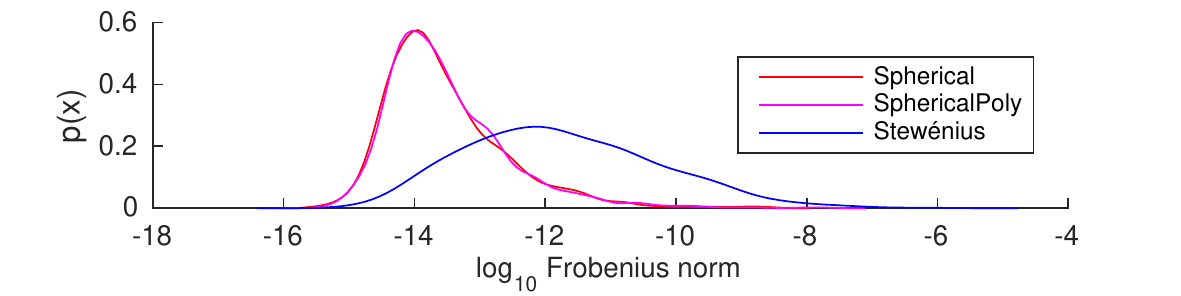}
    \includegraphics[width=6cm]{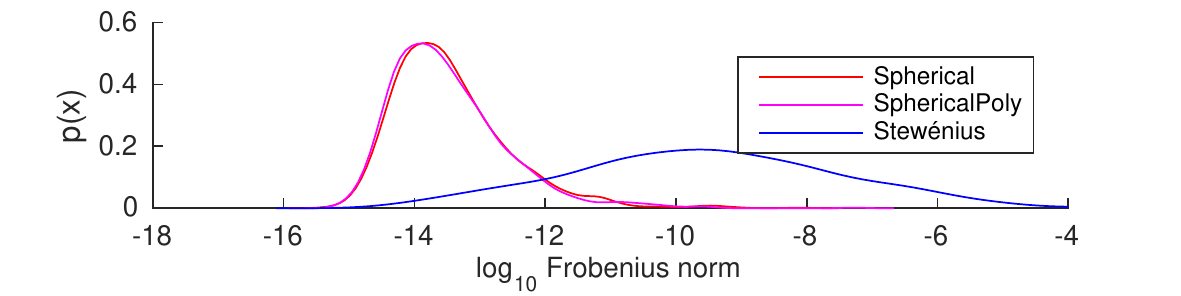}
    \caption{\label{fig:frobin}Kernel density plots for numerical error of minimal solvers with ideal observations and inward-facing cameras (\emph{top}) and outward-facing cameras (\emph{bottom}).}
\end{figure}

In this section we evaluate the speed and accuracy of our novel essential matrix solvers and compare them against the state-of-the-art five-point solution by \Stewenius~et al.~\cite{Stewénius2006284} for general camera motion.  We will refer to our action matrix solution from Section \ref{sec:actionmatrix} as \Spherical~and our polynomial solution from Section \ref{sec:polynomial} as \SphericalPoly, and we will refer to the five-point solution as \Stew.  We use the implementation of \Stew~provided in the OpenGV library \cite{6906582}.

\subsubsection{Random Problem Generation}  To make synthetic data for our tests, we generate random spherical motion problems using the following scheme.  First we generate a random rotation of the desired magnitude $\theta$ and calculate the first and second camera poses according to this relative rotation, so that both cameras lie on the unit sphere.  Then we randomly generate 3D points within a range of distances from the first camera; we use a distance range of $[0.25~0.75]$ for inward-facing cameras and $[4~8]$ for outward-facing cameras.  Each 3D point is projected into both cameras using a focal length of $600$ and Gaussian noise is added to the point observations with standard deviation of  $\sigma$ pixels.

\subsubsection{Timing} We calculated the average computation for our solvers over 10 000 randomly generated problems.  The testing was performed on a 2.6 GHz Intel Core i5 with optimized code written in C++.  \Spherical~and \SphericalPoly~takes 6.9 $\mu$s and 6.4 $\mu$s on average, respectively.  \Stew~takes 98 $\mu$s;  however, the implementation in OpenGV is not optimized for speed.

\subsubsection{Numerical Accuracy.}We tested the numerical accuracy of the solvers with ideal, zero-noise observations.  We generated 1000 random problems with a rotation of $\theta = 1$ degrees and $\sigma = 0$ pixels using both the inward- and outward-facing configuration.  We then ran each solver on the problem sets and calculated the Frobenius norm of the error between estimated and true essential matrix.  To test the minimal configuration, our solvers used three correspondences for estimation with a fourth for disambiguation and \Stew~used five corresponces for estimation with a sixth for disambiguation.  The results are plotted in Figure \ref{fig:frobin}.
    
Our spherical solvers are almost equivalent to each other in numerical accuracy and are two to four orders of magnitude more accurate than \Stew~for spherical camera motion estimation.

\begin{figure}
    \centering
    \includegraphics[width=12cm]{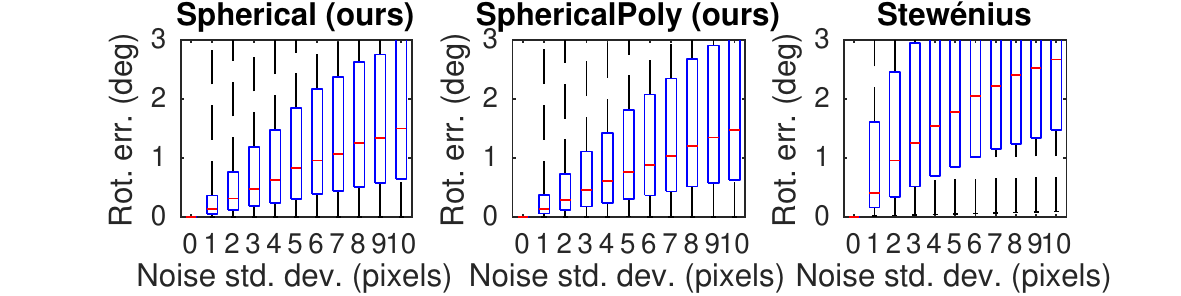}
    \includegraphics[width=12cm]{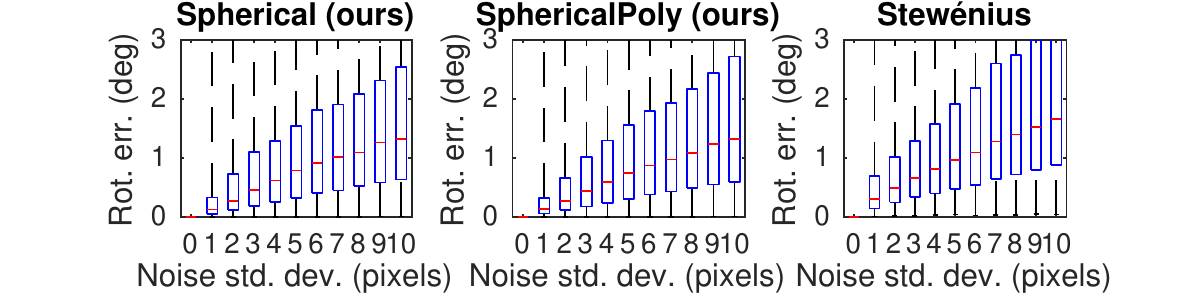}
    \caption{\label{fig:noisein}Box plots for angular error of minimal solvers with noisy observations and inward-facing cameras (\emph{top}) and outward-facing cameras (\emph{bottom}).}
\end{figure}

\subsubsection{Noise}We tested the robustness of the solvers to varying levels of added noise in the image observations.  We generated 1000 random problems with a rotation of $\theta = 1$ degrees and $\sigma = 0,1,\ldots,10$ pixels  using both the inward- and outward-facing configuration.  We then ran each solver on the problem sets and calculated the angular error $||\mathrm{log}_{SO(3)}( \Matrix{R}_{\mathrm{true}} \cdot  \Transpose{\Matrix{R}_{\mathrm{estimate}}} )||$.  For a fair comparison on noisy data, each solver used five correspondences for estimation.  The results are plotted in Figure \ref{fig:noisein}.

Again, both of our solvers are about equivalent in terms of accuracy and outperform \Stew~for spherical motion estimation with noisy correspondences.

\subsection{Structure-from-Motion Pipeline}

We tested the entire proposed structure-from-motion pipeline on several image sequences captured with both inward- and outward-facing configurations. We describe here the details of our implementation and show the resulting 3D reconstructions.

\subsubsection{Implementation Details}
In our experiments we apply the Oriented FAST and Robust BRIEF (ORB) feature detector \cite{6126544} and Kanade-Lucas-Tomasi (KLT) feature tracker \cite{lucas1981iterative,tomasi1991detection}.   The KLT tracker uses sub-pixel refinement which is especially helpful for a handheld, outward-facing camera where baseline the between images might be small relative to the scene depth.  We use an threshold of 2 pixels for both RANSAC inlier testing and the Huber cost function.

For efficiency, in our experiments we only detect loop closures with the first frame in the sequence.  We detect loop closures by iterating through the sequence backward from the last frame and stop when the number of inliers is below the threshold.  Feature matches are chosen as the nearest neighbor in Hamming distance between ORB descriptors.  A loop closure is accepted if it has at least 100 inliers.

The pipeline was implemented in C++ using OpenCV for image processing functions and Ceres \cite{ceres-solver} for non-linear optimization.  We set a minimum inverse depth of 0.01; points at this distance are essentially points at infinity. 

\subsubsection{Video Tests}We tested our system on several image sequences captured both indoors and outside with inward- and outward-facing camera configurations.  While the general motion of the test videos is circular, they were captured with a handheld camera and thus inevitably exhibit deviations from the circular path. A circular motion solver similar to \cite{5152255} failed on these sequences in our tests.

For the \emph{street} and \emph{bookshelf} sequences, we used a Sony $\alpha5100$ camera with 16mm lens.  For the \emph{face} sequence we used an Apple iPhone 5s.  Both devices were set to record 1080p video.

\begin{figure}[t]
    \centering
    \includegraphics[height=3cm]{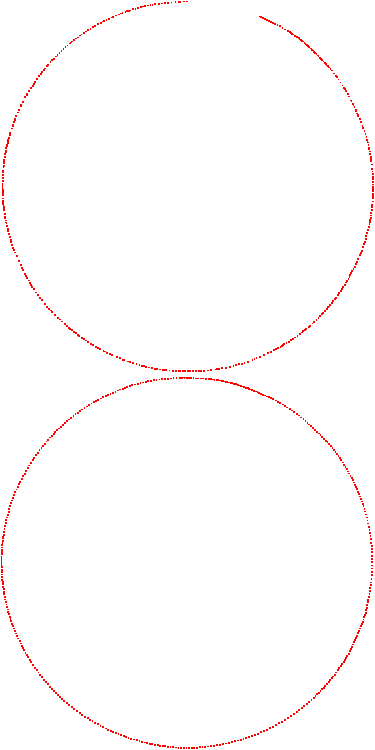}
    \includegraphics[height=3cm]{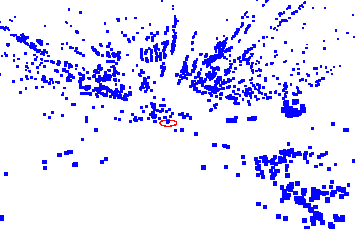}
    \hspace{0.5cm}
    \includegraphics[width=5cm]{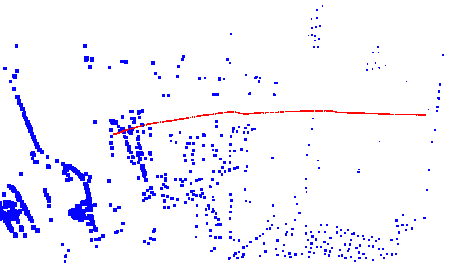}
    \caption{\emph{Top left}: Estimated camera centers from the \emph{street} sequence before loop closure. \emph{Bottom left}: Camera centers after rotation avergaging and bundle adjustment. \emph{Middle}: 3D reconstruction of the \emph{street} sequence.  The red dots are camera centers and the blue dots are reconstructed scene points.  \emph{Right}: 3D reconstruction of the \emph{bookshelf} sequence.}
    \label{fig:street3d}
\end{figure}

The \emph{street} sequence was captured in the middle of a neighborhood street corner.  We spun in a circle while holding the camera in our outstretched hands.  This sequence has a complete loop which is successfully detected by our system.  Figure \ref{fig:street3d} shows the reduction in drift after loop closure and a view of the complete 3D reconstruction.

The \emph{bookshelf} sequence was captured in an indoor office.  This sequence was also capture with outward-facing configuration but does not complete a full loop and has much closer objects than the \emph{street} sequence.  Figure \ref{fig:street3d} shows a view of the 3D reconstruction.

\begin{figure}[t]
    \centering
    \includegraphics[height=6cm]{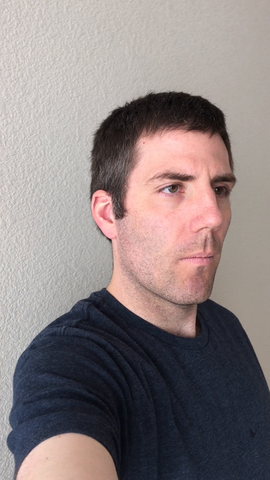}
    \hspace{0.25cm}
    \includegraphics[height=6cm]{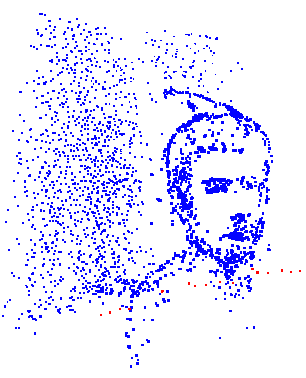}
    \includegraphics[height=6cm]{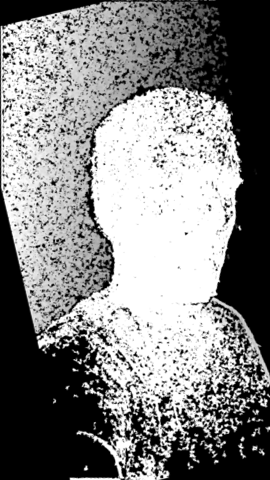}
    \caption{\emph{Left}: Image from the \emph{face} sequence.  \emph{Middle}: 3D reconstruction of the \emph{face} sequence.  The red dots are camera centers and the blue dots are reconstructed scene points. \emph{Right}: Disparity map from a rectified stereo pair from the \emph{face} sequence.}
    \label{fig:face3d}
\end{figure}

The \emph{face} sequence was captured by holding the iPhone in an outstretched hand with the lens pointed at the user's shoulder.  Rotating the arm produces inward-facing spherical motion which was used to capture a scan of the user's face.  Figure \ref{fig:face3d} shows a sample image and the 3D reconstruction.

\begin{figure}
    \centering
    \includegraphics[width=6cm]{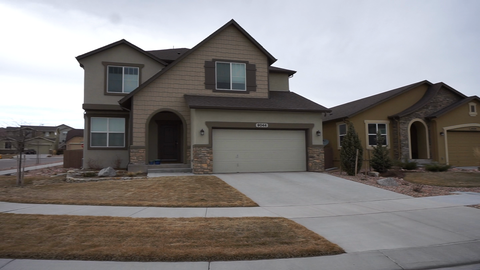}
    \includegraphics[width=6cm]{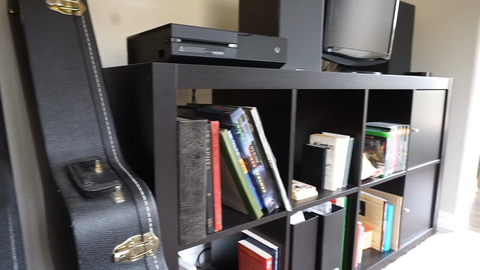}\\
    \includegraphics[width=6cm]{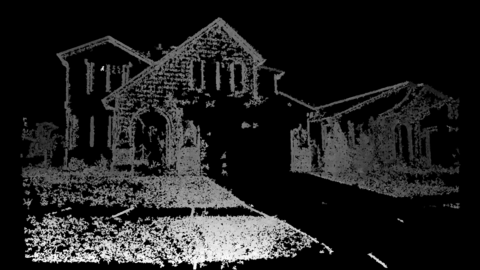}
    \includegraphics[width=6cm]{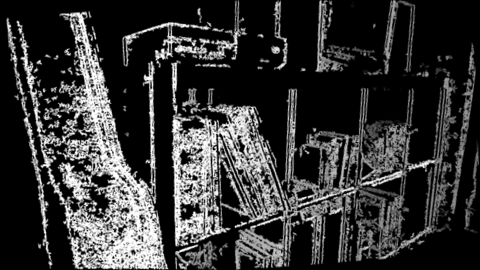}
    \caption{Disparity maps from from the \emph{street} and \emph{bookshelf} sequences.}
    \label{fig:stereo}
\end{figure}

To further illustrate the accuracy of these reconstructions, we selected image pairs from each sequence and performed stereo rectification using the recovered relative pose.  We used block matching to produce disparity maps as shown in Figures \ref{fig:face3d} and \ref{fig:stereo}.

\section{Conclusions and Future Work}
\label{sec:conclusions}

In this work we analyzed the geometry of spherical camera motion.  We introduced two solvers for the essential matrix arising from spherical motion.  The solvers require three point correspondences in the minimal case as opposed to the five needed for general motion, which reduces the number of hypothesese needed for random sample consensus.  The solvers are fast and exhibit better numerical accuracy and robustness to noise than the state-of-the-art.

By integrating these solvers into a structure-from-motion pipeline, we demonstrated that spherical motion greatly simplifies the problem by eliminating the need for translation estimation.  Despite the small baselines captured by a handheld camera, we found that accurate and large-scale reconstruction is possible using spherical structure-from-motion.

One limitation of the approach is the rigidness of the spherical motion constraint; deviations from spherical motion will cause the structure-from-motion pipeline to fail.  The system is less sensitive to deviations from the spherical constraint when the sphere's radius is small relative to scene depth; however, the precision of the 3D reconstruction also degrades as the radius-to-depth ratio decreases.  One possible way to alleviate this problem would be to increase the image resolution to make the parallax detectable again.

Future work includes more exploration of the potential applications of spherical structure-from-motion for user-friendly scene modeling and object scanning.

\section*{Acknowledgments}

This material is based upon work supported by the National Science Foundation under Grant No. 1464420.

\clearpage
\bibliographystyle{splncs03}
\bibliography{0475}
\end{document}